\documentclass{article}
\usepackage{spconf,amsmath,graphicx,amssymb,bm,booktabs,threeparttable,multirow,subfig,caption}
\usepackage{bibspacing}
\usepackage{color}

\newcommand{\todo}{\textcolor{red}{TODO}}

\newcommand{\seenclasses}{\textit{seen classes}}
\newcommand{\cut}[1]{}
\newcommand{\etal}{\textit{et al}.}
\newcommand{\ie}{\textit{i}.\textit{e}.}
\newcommand{\eg}{\textit{e}.\textit{g}.}
\newcommand{\etc}{\textit{etc}}

\title{Deep Zero-Shot Learning for Scene Sketch}

\name{Yao Xie$^{+}$\thanks{$^{+}$These authors contributed equally. ~~~$^{*}$ Corresponding author.} \qquad Peng Xu$^{+}$ \qquad Zhanyu Ma$^{*}$}

\address{\normalsize Pattern Recognition and Intelligent System Lab., Beijing University of Posts and Telecommunications, China.}

\begin{document}
\topmargin=0mm
%
\maketitle
\begin{abstract}

We introduce a novel problem of scene sketch zero-shot learning~(SSZSL), which is a challenging task, since~(i) different from photo, the gap between common semantic domain~(\eg, word vector) and sketch is too huge to exploit common semantic knowledge as the bridge
for knowledge transfer,
and~(ii) compared with single-object sketch, more expressive feature representation for scene sketch is required to accommodate its high-level of abstraction and complexity.
To overcome these challenges, we propose a deep embedding model for scene sketch zero-shot learning. In particular, we propose the augmented semantic vector to conduct domain alignment by fusing multi-modal semantic knowledge~(\eg, cartoon image, natural image, text description),
and adopt attention-based network for scene sketch feature learning. 
Moreover, we propose a novel distance metric to improve the similarity measure during testing.
Extensive experiments and ablation studies demonstrate the benefit of our sketch-specific design.
\cut{
Extensive experiments and ablation studies demonstrate that (i) state-of-the-art word vector based deep zero-shot learning models
specifically designed for photo fail to perform well on scene sketch,
(ii) comprehensive knowledge from other modalities can improve the domain alignment for scene sketch.}


\end{abstract}

\vspace{-0.1cm}
\section{Introduction}
\vspace{-0.1cm}
\label{sec:intro}

Sketch is abstract yet highly illustrative.
With the increasing popularity of touch-screen devices, more and more free-hand sketches are used for human-computer interaction (\eg, people can draw sketch as query to search specific shoe, chair, hand-bag~\cite{songjifei2017sketch}). The application conveniences of sketches have raised a flourish of sketch-related research, including recognition~\cite{schneider2014sketchFV}, sketch-based image retrieval~\cite{xu2016instance,songjifei2017sketch}, sketch hashing~\cite{xu2018sketchmate}, generation~\cite{hu2016now,Chen_2018_CVPR}, abstraction~\cite{riaz2018learning}, \etc.
However, most of the existing works focus on \textit{single-object sketches}~(\eg, apple, clock), leaving the \textit{scene sketches} under-studied.
Scene sketches are more abstract and complicated due to multiple objects and their interactions. In this paper,
we propose a novel problem of \textbf{scene sketch zero-shot learning}~(SSZSL), which is more challenging than scene sketch understanding~\cite{ye2016human,zou2018sketchyscene} and single-object sketch zero-shot learning~\cite{Yelamarthi2018ECCV}.



\textcolor{black}{Zero-shot learning methods rely on a labelled training set of \seenclasses~and the knowledge about how an \textit{unseen class} is semantically related to the seen classes. Seen and unseen classes are usually related in a high dimensional semantic knowledge domain, where the knowledge from seen classes ~can be transferred to unseen classes \cite{fu2015zero}.}
In previous computer vision works, word vector \cite{socher2013zero,Liu2017Generalized}, attribute vector \cite{ferrari2008learning,parikh2011relative,dong2017multi,wang2017attribute,li2018unsupervised} or text description \cite{reed2016learning} are widely studied as semantic knowledge, which is also used in the existing zero-shot learning methods engineered for photos.
 Different from photo, the gap between these semantic domains and sketch is too huge to exploit common semantic knowledge as the bridge for knowledge transfer.
Moreover, we aim to solve the scene sketch zero-shot classification.
Therefore, the main challenge to SSZSL is how to choose a reasonable semantic knowledge.
In this work, we exploit a novel semantic knowledge, termed as \textit{augmented semantic vector}, which can be obtained by fusing common semantic knowledge with the information from other modalities (\eg, cartoon image, natural image\textcolor{black}{, text description}).
Based on our augmented semantic vector, we propose a deep embedding model to solve scene sketch zero-shot learning,
in which we adopt visual feature space of scene sketch as the embedding space to alleviate hubness problem~\cite{zhang2017learning}.
Moreover, considering the high-level abstraction and complexity of scene sketch,
we use attention-based technique to obtain~discriminative feature representations for scene sketch. 
Most of the existing zero-shot learning (ZSL) methods use  either Euclidean distance~\cite{zhang2017learning} or cosine similarity~\cite{yelamarthi2018zero} as feature distance metric for testing.
 We define a new distance metric by combining Euclidean distance and cosine distance simultaneously, and achieved better evaluation results.
 \cut{
 Extensive experiments demonstrate that state-of-the-art ZSL models designed for photos fail to perform well for the scene sketch ZSL.}



\begin{figure}
  \centering
  \includegraphics[width= 0.3\textwidth]{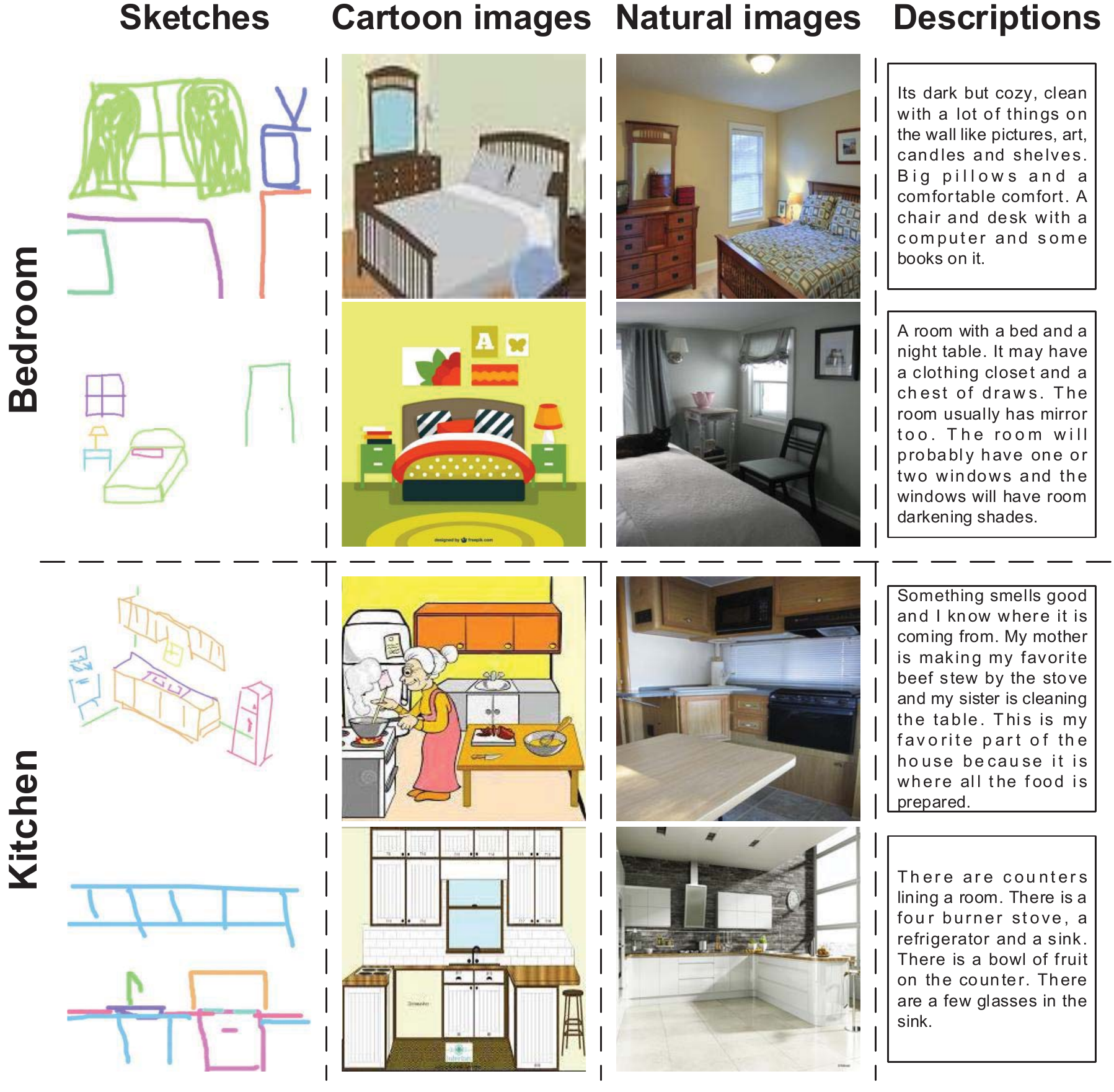} 
  \vspace{-0.3cm}
  \caption{Some samples from CMPlaces~\cite{castrejon2016learning}.
  } 
  \vspace{-0.5cm}
  \label{fig1} 
\end{figure}



\begin{figure*}
  \centering
  \includegraphics[width=16cm,height=4.6cm]{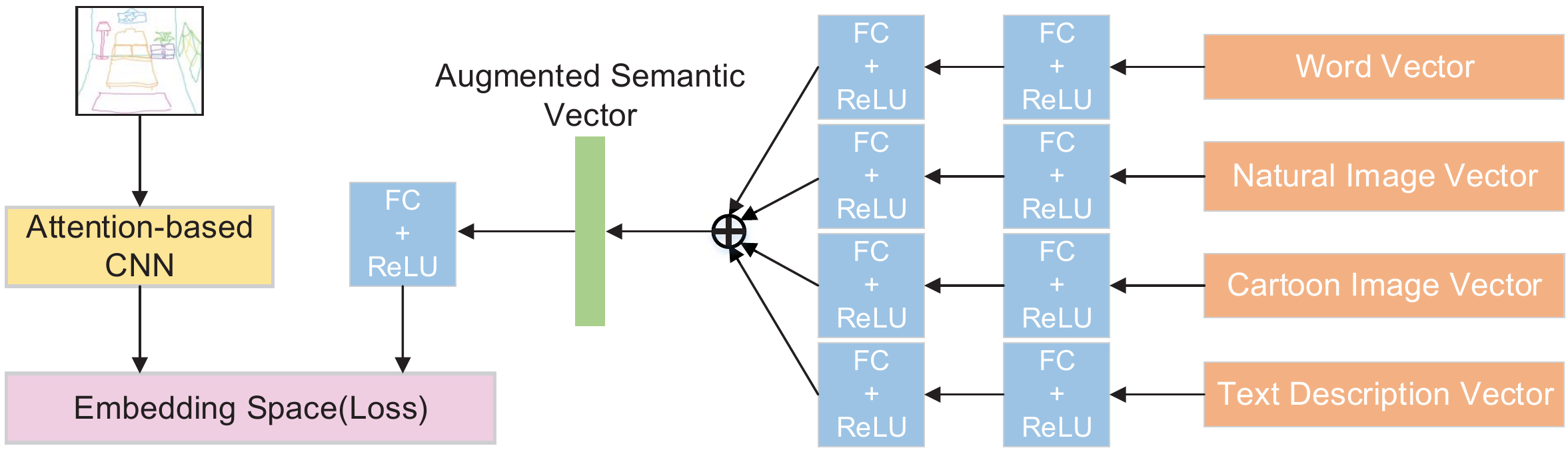}
  \vspace{-0.3cm}
  \caption{The network architecture of our SSZSL model.}
  \vspace{-0.5cm}
  \label{fig2} 
\end{figure*}


The contributions in this paper can be summarized as follows: (i)~To the best of our knowledge, this is the first time that zero-shot learning setting is defined in scene sketch classification task. We propose this novel problem, and illustrate its intrinsic traits.
(ii)~We propose a deep embedding model for scene sketch zero-shot learning~(SSZSL), which achieves a better performance than the state-of-the-art word vector based ZSL methods. Our superior performances effectively demonstrate the benefit of our sketch-specific design.
(iii)~Moreover, we define a new distance metric that performs better than conventional metric in SSZSL testing.


\vspace{-0.15cm}
\section{RELATED WORK}
\vspace{-0.1cm}
\label{sec:related}

Sketch can be used for human-computer interaction, thus sketch recognition has been the important research topic in recent years. Most of previous works focus on single-object sketch classification or retrieval~\cite{eitz2012humans,schneider2014sketchFV,xupeng2016Cross,zhong2018directional,xu2017cross}, leaving scene sketch classification under-studied. In particular, scene sketch zero-shot learning~(SSZSL) has not been studied to date.
To the best of our knowledge, only Ye \etal~\cite{ye2016human} have proposed a deep CNN model ``Scene-Net'' for scene sketch classification.

Existing ZSL methods engineered for scene photo mainly differ
in what semantic knowledge are used: typically either word vector \cite{socher2013zero}, attribute \cite{ferrari2008learning,parikh2011relative,zhu2018generalized} or text description \cite{reed2016learning}.
Sketch is different from photo. Due to the high-level abstraction, the domain gap between scene sketch and these common semantic knowledge is huge so that existing ZSL methods designed for photo fail to perform well on scene sketch. In this paper, considering the intrinsic traits of scene sketch, we propose a deep embedding model to solve SSZSL, in which we propose the augmented semantic vector as the \text{semantic knowledge} to conduct the knowledge transfer and domain alignment\cite{socher2013zero,li2019dual}.


\section{LEARNING A MODEL FOR SCENE SKETCH ZERO-SHOT CLASSIFICATION}
\label{sec:SSZSL}

\vspace{-0.1cm}
\subsection{Problem Formulation}
\vspace{-0.1cm}
\label{ssec:framework}

We now provide a formal deﬁnition of the scene sketch zero-shot learning~(SSZSL). Let
$\mathcal{S}_{tr} = \{({\bf S}_{i},{\bf x}_{i},{\bf y}^{u}_{i},{t}^{u}_{i})\}_{i=1}^{M}$
denote a labelled training set of $M$ training samples, where ${\bf x}_{i} \in{\mathbb R}^{J\times1}$ is the visual feature vector of the $i$-th training scene sketch ${\bf S}_{i}$.
${\bf y}^{u}_{i} \in {\mathbb R}^{L\times1}$
and ${t}^{u}_{i} \in\mathcal{T}_{tr}$ denote the semantic representation vector and class label of ${\bf S}_{i}$, which belongs to the $u$-th training class.

Given a new test sketch ${\bf S}_{j}$ with its feature visual vector ${\bf x}_{j}$, the goal of SSZSL is to predict a class label ${t}^{v}_{j}$ by learning a classifier $f$ : ${\bf x}_{j} \to {t}^{v}_{j}$, where ${t}^{v}_{j} \in\mathcal{T}_{te}$ is the class label of the $j$-th test instance~${\bf S}_{j}$ belonging to  $v$-th test class.
The training (seen) classes and test (unseen) classes are disjoint, \ie, $\mathcal{T}_{tr} \cap \mathcal{T}_{te} = \varnothing$. Note that each class label ${t}^{u}$ or ${t}^{v}$ is associated with a pre-defined semantic representation ${\bf y}^{u}$ or ${\bf y}^{v}$.


\vspace{-0.15cm}
\subsection{Deep Embedding Model}
\vspace{-0.1cm}
As is shown in Fig. \ref{fig2}, there are two branches in our model. The first branch is the visual feature extraction branch for scene sketch, composed of a attention-based network. It takes a scene sketch ${\bf S}_{i}$ as input and outputs a visual feature vector ${\bf x}_{i}$. To alleviate hubness problem, we will use this $J$-dimensional visual feature space as 
embedding space, where both the scene sketch and its corresponding semantic representation vector will be embedded. The second branch is the semantic embedding branch, which will embed the semantic knowledge into our embedding space to conduct the domain alignment. 
Considering the huge domain gap between common semantic domain~(\eg, word vector, text description) and high-level abstract scene sketch, we propose the augmented semantic vector for SSZSL.
In particular, our semantic embedding branch
\textcolor{black}{takes semantic representations from different modalities as input and it}
is implemented by three fully connected (FC) + Rectiﬁed Linear Unit (ReLU) layers with $\ell_2$ parameter regularization,
\textcolor{black}{where the first two FC+ReLU layers are used to obtain our augmented semantic vector~(see Sec.~\ref{ssec:augmented semantic vector}) and it will be embedded to the embedding space by the third FC+ReLU layer.}

Then the outputs of two branches are connected by a mean square error (MSE) loss which aims to minimize the difference between visual feature vector ${\bf x}_{i}$ and its embedded semantic vector in the visual feature space.
Our loss function is
\vspace{-0.2cm}
\begin{equation}
\vspace{-0.1cm}
\label{eq:loss}
J({\bf W}) = \frac{1}{M}\sum_{i = 0} ^M\|{\bf x}_{i} - \Phi(\textcolor{black}{{\bf y}^{u}_{i}}\!:\!{\bf W}) \|_2^2 \\
+ \lambda\|{\bf W}\|_2^2,
\end{equation}
where ${\bf W}$ contains the weights of the semantic embedding branch. $\lambda$ is the hyperparameter to weight the strength of the $\ell_2$ parameter regularization loss against the MSE loss.
$\Phi(\cdot)$ denotes the feature extraction by our semantic branch.

After training, 
to predict label $t^{v}_{j}$,
we can calculate the distance between the embedded semantic vectors and ${\bf x}_{j}$
as
\vspace{-0.2cm}
\begin{equation}
\vspace{-0.2cm}
t^{v}_{j} \gets v = \arg \min\limits_{v} {D}(\Phi({\bf y}^{v}), {\bf x}_{j}),
\end{equation}
where
$D$
is a distance metric function defined by us~(see Sec.~\ref{sec:our_metric}),
and ${\bf y}^{v}$ is the embedded semantic vector 
of the $v$-$th$ unseen class.

\vspace{-0.1cm}
\subsection{Augmented Semantic Vector for Scene Sketch}
\vspace{-0.1cm}
\label{ssec:augmented semantic vector}

In this work, we exploit a novel semantic knowledge for SSZSL, termed as augmented semantic vector.
As illustrated in Fig.~\ref{fig2}, our semantic embedding branch takes in semantic representations from different modalities~(\ie, word vector (W), natural image vector (I), cartoon image vector~(C), text description vector~(T)), and after the first two FC+ReLU layers, it outputs the angmented semantic vector~$\textcolor{black}{\tilde{\bf y}_{i}^{u}}$ by element-wise addition and non-linear fusing
\vspace{-0.2cm}
\begin{equation}
\vspace{-0.2cm}
\begin{split}
\tilde{\bf y}_{i}^{u} = & \Phi^{W}(({\bf y}_{i}^{u})^{W}) + \Phi^{C}(({\bf y}_{i}^{u})^{C})+ \\
& \Phi^{I}(({\bf y}_{i}^{u})^{I})+ \Phi^{T}(({\bf y}_{i}^{u})^{T}),
\end{split}
\end{equation}
where $\Phi^{W}$, $\Phi^{C}$, $\Phi^{I}$, and $\Phi^{T}$ denote the mapping of the first two FC layers of our semantic branch~(See Fig.~\ref{fig2}).
$({\bf y}_{i}^{u})^{W}$, $({\bf y}_{i}^{u})^{C}$, $({\bf y}_{i}^{u})^{I}$, and $({\bf y}_{i}^{u})^{T}$ denote the representation vectors from four modalities.
To obtain natural image vector, we train a deep classifier network based on natural images and calculate a vector representation for each sample class by averaging the deep features by category.
For cartoon image vector and text description vector, the similar process is used. For word vector, we adopt the word2vec model (based on model library in Gensim), which was trained with over $8,000,000$ text documents from Wiki-pedia, to represent each class (including seen and unseen classes).

\vspace{-0.1cm}
\subsection{Distance Metric}
\vspace{-0.1cm}
\label{sec:our_metric}

\begin{figure}
  \centering
  \includegraphics[width= 0.25\textwidth]{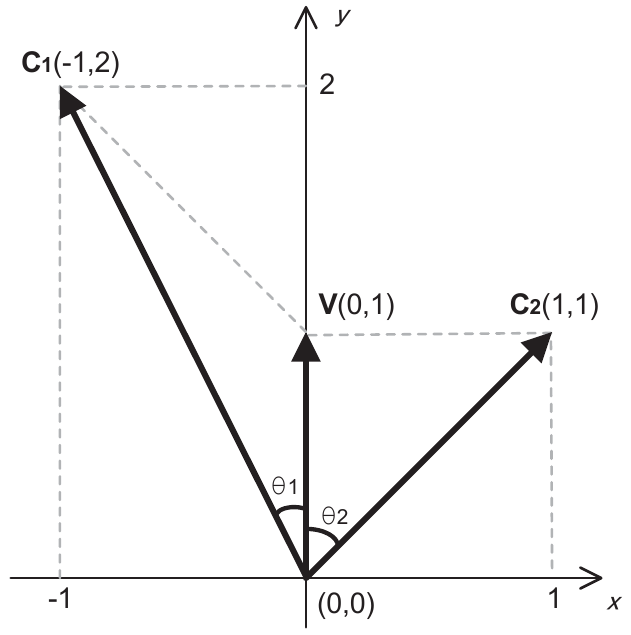}
  \vspace{-0.3cm}
  \caption{\textcolor{black}{Illustration for distance metric.}} 
  \label{fig3} 
 \vspace{-0.5cm}
\end{figure}

Euclidean distance is a frequently-used distance metric,
however it
will fail for
the case shown in Fig. \ref{fig3}.
$\bm{c}_{1}$ and $\bm{c}_{2}$ represent the embedded semantic representation vectors corresponding to two categories. $\bm{v}$ is a visual feature vector belonging to $\bm{c}_{1}$, where $\|\bm{v}-\bm{c}_{1}\|^{2}>\|\bm{v}-\bm{c}_{2}\|^{2}$ and
$\theta_{1} < \theta_{2}$.
If only using Euclidean distance, $\bm{v}$ will be classified to $\bm{c}_{2}$.
However, if only use Cosine distance, $\bm{v}$ will be classified to $\bm{c}_{1}$.
Therefore we propose a new distance metric termed as Euclidean Cosine (EC) distance to alleviate this problem.
Our EC distance metric is defined as follows:
\vspace{-0.2cm}
\begin{equation}
\vspace{-0.2cm}
\label{eq:metric}
\left\{ \begin{array}{ll}
    D(\bm{\alpha}, \bm{\beta}) = (1- \eta cos\langle\bm{\alpha},\bm{\beta}\rangle) \|\bm{\alpha} - \bm{ \beta}\|_2^2\\
    cos\langle\bm{\alpha}, \bm{\beta}\rangle \ = \ \frac{\bm{\alpha} \cdot \bm{\beta}}{\|\bm{\alpha}\| \cdot \|\bm{\beta}\|}
    \end{array} \right.,
\end{equation}
where $\bm{\alpha}$ and $\bm{\beta}$ are vectors, and
$\eta (0\!\!\le\!\!\eta\!\!\le \!\!1)$ is a weighting coefﬁcient that controls the importance of cosine distance.
Assuming that $\eta$ is $0.9$,
in Fig.~\ref{fig3}, we can obtain 
$D(\bm{v},\bm{c_{1}})<D(\bm{v},\bm{c_{2}})$, where $D(\bm{v},\bm{c_{1}})$ and $D(\bm{v},\bm{c_{2}})$ are approximately $0.2758$ and $0.3636$, respectively.

\section{EXPERIMENT}
\vspace{-0.1cm}
\label{sec:experiment}


\begin{table*}[t]
\small
\begin{center}
\caption{SSZSL (Top1/Top5) accuracy (\%) comparison with state-of-the-art ZSL models on CMPlaces.}
\vspace{-0.3cm}
\label{tab:comparison_with_sota}
\resizebox{\textwidth}{!}{
\begin{tabular}{c|ccccc|cccccc}
\hline
\multirow{2}{*}{Model}&
    \multicolumn{5}{c|}{\textcolor{black}{Default Metric}}&\multicolumn{5}{c}{EC Distance Metric}\\
    \cline{2-6}\cline{7-11}
    &W&C&I&T&W+C+I+T&W&C&I&T&W+C+I+T\\
\hline
\hline
\bf DEM ($\bf V \to S$) \cite{zhang2017learning}&23.5/39.9&30.9/70.5&26.8/61.4&38.2/77.2&38.8/78.6&30.2/58.3&34.8/72.8&27.3/62.3&39.3/77.9&40.4/79.3  \\
\bf DEM ($\bf S \to V$) \cite{zhang2017learning}&23.9/54.4&41.7/79.6&36.9/78.3&41.3/82.1&47.7/84.1&30.9/67.5&44.6/81.4&37.7/80.7&42.5/82.6&48.1/84.7  \\
\bf SAE \cite{kodirov2017semantic}&25.9/58.9&34.1/72.5&27.5/71.1&33.1/76.7&45.5/84.1&30.7/66.2&34.5/73.8&29.7/73.6&35.3/77.1&48.7/86.2  \\
\bf f-CLSWGAN \cite{xian2018feature}&18.1/44.1&39.9/73.3&32.5/70.5&37.3/70.9&48.7/83.4&-&-&-&-&- \\
 \bf RELATION NET \cite{yang2018learning} &30.6/66.1&40.3/76.4&35.8/74.5&40.5/79.1&47.6/82.5&-&-&-&-&- \\
\hline
\bf Ours&30.4/61.5&45.2/80.5&38.9/81.6&43.4/82.9&52.1/85.7&\bf 35.6/70.2& \bf 46.9/81.8&\bf 40.1/82.8&\bf 45.2/83.7&\bf 54.0/86.9\\
\hline
\end{tabular}
}
\end{center}
\vspace{-0.6cm}
\end{table*}

\vspace{-0.1cm}
\subsection{Dataset}
\vspace{-0.1cm}
\label{ssec:dataset}

In the following experiments,
CMPlaces~\cite{castrejon2016learning} servers as the benchmark for our SSZSL experiment, which contains hundreds of natural scene categories across five modalities, including natural images, sketches, cartoons images, text descriptions, and spatial text images\footnote{Spatial text images are not available in current online CMPlaces dataset.}.
Some samples of CMPlaces are illustrated in Fig.~\ref{fig1}, and we can observe that the scene sketches are abstract and have rich information.
All the following zero-shot experiments are performed for scene sketch, while natural images~($\bf I$), cartoons images~($\bf C$) and text descriptions ($\bf T$)
are used to obtain our augmented semantic vector.
Ignoring some classes that are too messy to recognize, we selected approximate $15, 000$ scene sketches from $174$ classes.
In particular, $144$ classes were used as seen classes and the remaining $30$ classes were used as unseen classes.

\vspace{-0.1cm}
\subsection{Experimental Settings}
\vspace{-0.1cm}
\label{ssec:experimentalsettings}
All our experiments are implemented in PyTorch
, and on a single GEFORCE GTX $1080$ Ti GPU card.

\textbf{Visual Feature Extraction Branch}
We use ImageNet pretrained SE-Resnet-50~\cite{hu2017squeeze} as our attention-based CNN model to conduct visual feature extracting for scene sketches, and only modify its fully connected layers to
fine-tune it on the scene sketches from $144$~seen classes.
The output dimension of our visual feature extraction branch is $2048$.
We use SGD optimizer with a initial learning rate $lr$ of $0.001$, momentum of $0.9$ and a mini-batch size of $16$.
We decrease $lr$ by $0.9$ every epoch and terminate the optimization after 30 epochs.
All input sketches or images are resized to $3 \times 224 \times 224$. The loss weighting factor $\lambda$ in Eq.~\ref{eq:loss} and hyperparameter $\eta$ in Eq.~\ref{eq:metric} are experimentally set to $0.0005$ and $0.9$, respectively.


\cut{
\begin{table}[tbp]
\small
\begin{center}
\caption{SSZSL (Top1/Top5) accuracy (\%) comparison with state-of-the-art ZSL models on CMPlaces.}
\label{tab:comparison_with_sota}
\resizebox{1.0\columnwidth}{!}{
\begin{tabular}{c c c c c c}
\hline
{\bf Model}&{\bf W}&{\bf C}&{\bf I}&{\bf T}&{\bf W+C+I+T}  \\
\hline
\hline
\bf DEM ($\bf V \to S$) \cite{zhang2017learning}&23.5/39.9&30.9/70.5&27.3/61.4&38.2/77.2&38.8/78.6  \\
\bf DEM ($\bf S \to V$) \cite{zhang2017learning}&23.9/54.4&41.7/79.6&38.9/80.7&41.3/82.1&47.7/84.8  \\
\bf SAE \cite{kodirov2017semantic}&29.3/63.7&34.5/73.6&29.8/73.4&35.7/76.8&48.4/86.1  \\
\bf f-CLSWGAN \cite{xian2018feature}&18.1/44.1&39.9/73.3&32.5/70.5&37.3/70.9&48.7/83.4 \\
 \bf RELATION NET \cite{yang2018learning} &30.6/66.1&40.3/76.4&35.8/74.5&40.5/79.1&47.6/82.5 \\
\hline
\bf Ours&\bf 35.6/70.2& \bf 46.9/81.8&\bf 40.1/82.8&\bf 45.2/83.7&\bf 54.0/86.9 \\
\hline
\end{tabular}
}
\end{center}
\end{table}}

\begin{table}[tbp]
\small
\begin{center}
\caption{Ablation study for our proposed model: SSZSL (Top1/Top5) accuracy (\%) on CMPlaces.}
\vspace{-0.3cm}
\label{tab:tab1}
\resizebox{1.0\columnwidth}{!}{
\begin{tabular}{c|cc|cc}
\hline
\multirow{2}{*}{Semantic Representation}&
    \multicolumn{2}{c|}{EC Distance}&\multicolumn{2}{c}{Euclidean Distance}\\
    \cline{2-3}\cline{4-5}
    &$\bf S \to V$ & $\bf V \to S$ & $\bf S \to V$ & $\bf V \to S$\\
\hline
\hline
    W&35.6/70.2&33.6/61.3&30.4/61.5&25.1/41.9 \\
    C&{\bf 46.9}/81.8&36.5/74.6&45.2/80.5&32.6/71.2 \\
    I&40.1/82.8&29.4/63.9&38.9/81.6&28.9/63.2 \\
    T&45.2/\bf 83.7&42.5/78.9&43.4/82.9&41.4/77.9 \\
    \hline
    \hline
    C+T&\bf 51.1/\bf 86.3&44.8/81.0&49.6/84.6&42.7/78.5\\
    I+T&49.4/85.6&44.6/81.1&48.3/84.9&40.6/80.3 \\
    I+C&47.9/84.6&36.7/78.0&46.1/83.2&36.5/77.8 \\
    W+C&47.5/82.4&41.8/75.6&44.6/80.9&34.5/68.7 \\
    W+T&46.7/84.1&44.4/79.7&45.1/82.7&40.4/74.7 \\
    W+I&45.1/83.4&31.9/68.4&43.3/80.9&29.7/64.2 \\
    \hline
    \hline
    C+I+T&\bf 52.8/\bf 86.4&43.2/81.4&51.7/85.6&42.8/80.3 \\
    W+C+I&49.0/84.6&38.0/78.8&48.6/83.2&36.6/77.8 \\
    W+C+T&51.7/86.2&44.7/80.4&50.5/85.2&41.0/77.5 \\
    W+I+T&50.4/86.1&42.3/79.7&49.6/84.9&41.6/78.9 \\
    \hline
    \hline
    W+C+I+T&\bf 54.0/\bf 86.9&43.6/81.4&52.1/85.7&42.4/79.8 \\
\hline
\end{tabular}
}
\end{center}
\vspace{-0.6cm}
\end{table}

\cut{
\textbf{SSZSL Model Implementation.}

We use ImageNet based pretrained SE-Resnet-50, and only modify its fully connected layers to
fine-tune it on the scene sketches from $144$~seen classes.
outputs a visual feature space with dimension $J = 2048$.  \todo optimizer.

}

\textbf{Semantic Embedding Branch}
In our semantic embedding branch, the outputs of three FC layers are set to $512$, $1024$, and $2048$, respectively.
Adam is used to optimise our semantic embedding branch with a initial learning rate of $0.0001$ and a mini-batch size of $256$.

\cut{
\textbf{Semantic Vector.\todo}
As shown in Fig. \ref{img2}, the CNN classifier (CNN net + two FC layers) is trained by natural /cartoon images, and the LSTM classifier (LSTM net with two hidden layers + one FC layer) is trained by text descriptions. The outputs of FC layers in CNN classifier are set to $512$ and $174$, respectively. The outputs of Fc layer in LSTM classifier is set to $174$. The word embedding size and the number of LSTM unit are both $300$. SGD with momentum is used to optimise model with a learning rate of $0.001$ and a minibatch size of $16$.
\todo


In this paper, we adopt the word2vec model (based on model library in Gensim), which was trained with over $8,000,000$ text documents from Wiki-pedia, to represent each class (including seen and unseen classes)
by a semantic vector~(equally set as $174$-dimensional).
}



\textbf{Competitors.}
As aforementioned state, SSZSL is a novel problem, thus there is no existing methods can perform as our baselines.
Therefore, we have to compare with state-of-the-art ZSL methods, including
 DEM~\cite{zhang2017learning},
SAE~\cite{kodirov2017semantic},
f-CLSWGAN~\cite{xian2018feature}, and
RELATION NET~\cite{yang2018learning}.



\cut{

\begin{table}[tbp]
\small
\begin{center}
\caption{SSZSL (Top1/Top5) accuracy (\%) on different distance metrics.}
\label{table:edgemap_benchmark}
\resizebox{0.95\columnwidth}{!}{
\begin{tabular}{c c c c c}
\hline
{\bf Model}&{\bf W}&{\bf C}&{\bf I}&{\bf T}  \\
\hline
\hline
\bf Euclidean distance&30.4/61.5&45.2/80.5&37.7/81.6&43.4/82.9  \\
\bf EC distance&\bf 35.3/71.1& \bf 46.9/81.8&\bf 40.1/82.8&\bf 45.1/83.7 \\
\hline
\end{tabular}
}
\end{center}
\end{table}

}

\vspace{-0.1cm}
\subsection{Results and Discussion}
\vspace{-0.1cm}
\label{ssec:Results}

\cut{
\todo Table~\ref{tab:tab1} reports the accuracy of SSZSL on word vector, augmented semantic vector.
According to the choice of embedding space, the results in Table 1 can be divided into two groups: visual feature embedding space ($S \to V$) and semantic embedding space ($V \to S$). W stands for word vector and C, I, T represent three augmented semantic vectors (Sec. 3.2). From Table. 1, We can make the following observations: (1) Our proposed three augmented semantic vectors achieve a better performance than word vector on SSZSL; (2) For single semantic space, cartoon images (C) and text descriptions (T) augmented semantic vectors in $S \to V$ achieve the best results on Top1 and Top5, respectively; For fusion semantic space, each of them in $S \to V$ outperforms its constituents and C+T in $S \to V$ achieve the best result on both Top1 and Top5; (3) The performance of our model in $S \to V$ is better than that in $V \to S$, and the augmented semantic vector is more suitable for our model in $S \to V$.   \todo
}

First of all, we compare our proposed model with the state-of-the-art ZSL models on CMPlaces dataset, as illustrated in Tab.~\ref{tab:comparison_with_sota}.
For a fair comparison, we use Resnet-$50$ as their visual feature extractor.
These selected competitors have performed well using word vector~(W) as semantic input~\cite{zhang2017learning,kodirov2017semantic,xian2018feature,yang2018learning}, thus we also evaluate them based on word vector.
Moreover, in order to demonstrate the importance of semantic knowledge for SSZSL, other semantic knowledge~(\ie, natural image vector (I), cartoon image vector~(C), text description vector~(T)), and their fused vector~(W+C+I+T)~are also evaluated as semantic input.
In our experiments, we find that our proposed distance metric outperforms both of Euclidean distance and cosine distance for SSZSL testing, and cosine distance performs really poor. Therefore,
if our proposed metric can be applied to these  ZSL baselines in Tab.~\ref{tab:comparison_with_sota}, they are also calculated on our proposed metric.
In Tab.~\ref{tab:comparison_with_sota}, we can observe that: (i) Our proposed model outperforms all the competitors by a large margin based on different semantic knowledge.
This is benefit from that we choose sketch feature space as a reasonable embedding space and our attention-based network achieves better feature representation.
(ii) When using the augmented semantic vector based on four modalities, our model obtains a obvious performance improvement,
and our augmented semantic vector also improves the performances of all the selected baselines.
This demonstrates the superiority of our sketch-specific augmented semantic vector.

\cut{
we can observe from Table 2 that our proposed method achieves the best results in four semantic spaces. We also make comparison to the state-of-the-art ZSL models on CMPlaces data set, which is shown in Table 2. We can observe from Table 2 that our proposed method achieves the best results in four semantic spaces. Table 3 report the difference between our proposed distance metric formula (EC distance) and Euclidean distance. We can observe from Table 3 that the result of our proposed EC distance is higher than that of Euclidean distance.}

The results of our ablation study are reported in Tab.~\ref{tab:tab1}.
As aforementioned state, our semantic embedding branch is scalable that can be adaptive to combinations of semantic vectors from different modalities, thus we evaluate our proposed model based on these combinations. In Tab.~\ref{tab:tab1}, we can make following observations:
(i) For SSZSL, our augmented semantic vector performs better than single-modal semantic vector.
(ii) Choosing a reasonable embedding space is important. Using sketch visual feature space as embedding space ($S \to V$) obtains better performance than using semantic space as embedding space ($V \to S$). This interesting phenomenon can be explained by the hubness issue discussed in~\cite{zhang2017learning}.
(iii) Our proposed EC distance outperforms Euclidean distance for zero-shot testing on CMPlaces.
(iv) When using single-modal semantic vector for our model,
cartoon image vector outperforms word vector by a clear margin~($46.9\%$ vs. $35.6\%$), since the domain gap between cartoon images and sketch is smaller.

\vspace{-1cm}
\section{CONCLUSION}
\vspace{-0.5cm}
\label{sec:conclusion}
In this paper, we introduce a novel problem of scene sketch zero-shot learning (SSZSL). We have proposed a deep embedding model for scene sketch zero-shot learning. The model differs from existing ZSL model in that we propose the augmented semantic vector to conduct domain alignment by fusing multi-modal semantic knowledge, and adopt attention-based network for scene sketch feature learning. What's more, a new distance metric is used instead of Euclidean distance. Experimental results on CMPlaces validated the effectiveness of the proposed method.



\textbf{Acknowledgement}
This work was supported in part by the National Natural Science Foundation of China (NSFC) under Grant No. 61773701, in part by the Beijing Nova Program No. Z171100001117049, in part by the Beijing Nova Program Interdisciplinary Cooperation Project No. Z181100006218137, and in part by BUPT Excellent PhD Student Foundation CX2017307 and BUPT-SICE Excellent Graduate Student Innovation Foundation (2016).

\bibliographystyle{IEEEbib}
\setlength{\itemsep}{-2mm}
\bibliography{Template}

\end{document}